\pdfoutput=1
\documentclass{article}

\PassOptionsToPackage{numbers, compress}{natbib}

\usepackage[preprint]{neurips_2023}

\usepackage[utf8]{inputenc} 
\usepackage[T1]{fontenc}    
\usepackage{hyperref}       
\usepackage{url}            
\usepackage{booktabs}       
\usepackage{amsfonts}       
\usepackage{nicefrac}       
\usepackage{microtype}      
\usepackage{xcolor}         
\usepackage{graphicx}
\usepackage{colortbl}
\usepackage{multirow}
\usepackage{amsmath}
\DeclareMathAlphabet\mathbfcal{OMS}{cmsy}{b}{n}
\usepackage{centernot}
\usepackage[linesnumbered,algoruled,boxed,noend]{algorithm2e}
\usepackage{bm}
\usepackage{enumitem}
\usepackage{booktabs, tabularx}

\newcommand{\ie}{\textit{i.e.}} 
\newcommand{\eg}{\textit{e.g.}} 
\definecolor{lightgreen}{RGB}{201,242,155}
\definecolor{darkgreen}{RGB}{0,170,136}
\definecolor{na}{gray}{0.9}

\usepackage{amsmath,amsfonts,bm}

\def\eqref#1{equation~\ref{#1}}

\def\1{\bm{1}}

\DeclareMathAlphabet{\mathsfit}{\encodingdefault}{\sfdefault}{m}{sl}
\SetMathAlphabet{\mathsfit}{bold}{\encodingdefault}{\sfdefault}{bx}{n}

\SetKwInput{KwInput}{Input}
\SetKwInput{KwOutput}{Output}
\SetEndCharOfAlgoLine{}

\newcommand{\nop}[1]{}

\title{Chain of Natural Language Inference for Reducing Large Language Model Ungrounded Hallucinations}

\author{Deren Lei\thanks{\enspace Equal contributions.}, \qquad Yaxi Li\footnotemark[1],\qquad Mengya (Mia) Hu\footnotemark[1],\qquad Mingyu Wang\footnotemark[1],\qquad Vincent Yun, \\ \textbf{Emily Ching,\qquad Eslam Kamal}\\
Microsoft Responsible AI\\
\{derenlei, yaxi.li, humia, mwang, xi.yun, yuetc, eskam\}microsoft.com\\
}

\begin{document}

\maketitle

\begin{abstract}
Large language models (LLMs) can generate fluent natural language texts when given relevant documents as background context. This ability has attracted considerable interest in developing industry applications of LLMs. However, LLMs are prone to generate hallucinations that are not supported by the provided sources. In this paper, we propose a hierarchical framework to detect and mitigate such ungrounded hallucination. Our framework uses Chain of Natural Language Inference (CoNLI) for hallucination detection and hallucination reduction via post-editing. Our approach achieves state-of-the-art performance on hallucination detection and enhances text quality through rewrite, using LLMs without any fine-tuning or domain-specific prompt engineering. We show that this simple plug-and-play framework can serve as an effective choice for hallucination detection and reduction, achieving competitive performance across various contexts. \footnote{\small \url{https://github.com/microsoft/CoNLI_hallucination}}
\end{abstract}

\section{Introduction}
Large Language models, known for their remarkable capabilities in natural language generation (NLG) \cite{touvron2023llama, OpenAI2023GPT4TR, chowdhery2022palm}, have attracted unprecedented interest from the public. These models serve as the foundation for a wide array of business applications (\eg Bing.com, ChatGPT, and Github Copilot). A common characteristic of such applications is their reliance on LLMs for text-to-text generation, often necessitating that the generated responses maintain factual consistency with the source text. Therefore, ensuring factual consistency is a critical challenge when evaluating the quality of generated responses \cite{maynezetal2020faithfulness, nanetal2021entity}. However, generating hallucination that diverges from the source text is a well-known phenomenon of LLMs. These hallucinations can be attributed to various factors, such as long input context \cite{liu2023lost}, irrelevant context distraction \cite{Shi2023LargeLM}, or complicated reasoning \cite{wei2022chain}. This phenomenon poses a significant challenge to the reliability of LLMs in real-world applications. 

Hallucination is commonly categorized as: \textit{context-related hallucination}, refers to hallucination where generated response contradicts commonsense; \textit{self-conflicting hallucination}, where generated response sentences conflict with each other (\eg numerical multi-step reasoning failed at a particular step \cite{chen2022program, zhang2023language}); and \textit{ungrounded hallucination}, where generated sentences conflict with the source text \cite{zhang2023siren} without assessing response coherence. Self-conflicting hallucination is more solution-dependent and behaves differently per downstream task. To generically enhance the reliability of LLM responses, our investigation focuses on reducing ungrounded hallucination, irrespective of the upstream task. We define alignment level with source as \textbf{groundedness} of LLM output.

Numerous existing works have concentrated on evaluating the groundedness of generated texts by developing classification \cite{zhou2021detecting, kryscinski2020evaluating, Zha2023AlignScoreEF} or ranking \cite{falkeetal2019ranking} models. While these detection models are useful in assessing groundedness, they provide limited utility in terms of rewriting and enhancing groundedness of a given LLM response.

Recent studies have explored methods for enhancing groundedness of LLM responses, including changing decoding strategy \cite{chuang2023dola}, inference-time self-critique \cite{press2022measuring, manakul2023selfcheckgpt}, multi-agent debate \cite{du2023improving}, and user-specified retrieval corpus \cite{gaoetal2023rarr}. In contrast, we study how to reduce hallucination when the user does not have full control over the LLM model or cannot leverage additional external knowledge. We propose a generic post-edit approach, named\textbf{ Chain of Natural Language Inference (CoNLI)}. In this framework, users are only required to bring their own text-to-text inputs/outputs and an LLM API endpoint. It will (1) select sentences as claims, (2) detect hallucination hierarchically with sentence-level and entity-level detectors (with a given entity detection model) by asking LLM to solve a sequence of natural language inference problems, and (3) leverage detection response in hallucination mitigator to get a refined response. We conducted experiments with CoNLI on text abstractive summarization and grounded question-answering scenarios with the latest hallucination benchmarks, both synthetic-generated and human-annotated. Our proposed approach demonstrates hallucination detection improvement against the latest solutions. Furthermore, the final refined responses show improvements over the initial provided response on various NLG evaluation metrics and groundedness metrics. Our interpretable and high-quality hallucination detection and reduction framework utilizes domain-agnostic few shots with simple post-editing techniques that prioritize the preservation of the original raw responses. We claim that our proposed framework is a generic solution that can potentially benefit various LLM-based business applications.


\section{Problem and preliminaries}
Previous research has encompassed different problem definitions and terminologies, often blending together aspects such as judging the correctness of text in various contexts, including free-text generation and text-to-text generation. Terminologies such as hallucination \cite{zhou2021detecting, manakul2023selfcheckgpt,li2023helma}, attribution \cite{gaoetal2023rarr}, factual consistency \cite{Zha2023AlignScoreEF,wang2020asking}, factuality \cite{goyal2020}, factual correctness \cite{zhang2020opt}, faithfulness \cite{maynezetal2020faithfulness, dong2022Faithful}, and truthfulness\cite{zheng2023truth}. In contrast, our focus exclusively centers on ungrounded hallucination, a phenomenon prevalent in text-to-text generation scenarios. It refers to any erroneous text generated by models that either conflict with or cannot be verified against the source texts. 

For text-to-text generation, we denote the input \textit{source text} as $X$ and the output \textit{raw response} as $Y_\text{raw}$, where $X$ and $Y_\text{raw}$, represented as $X = \{x_1, x_2,...,x_m\}$ and $Y_\text{raw} = \{y_1, y_2,...,y_n\}$ respectively, comprise one or more sentences. The generation can thus be denoted as:
\begin{equation}
    \mathbfcal{F}: X \rightarrow Y_\text{raw}
\end{equation}

In contemporary approaches, $\mathbfcal{F}(\cdot)$ is primarily powered by the language model. We say $Y_\text{raw}$ is grounded by $X$ if a generic reader would affirm the statement "According to $X$, $Y_\text{raw}$ is true" \cite{rashkin2021measuring}. Conversely, $Y_\text{raw}$ is hallucinated with respect to $X$ if it conflicts with or cannot be verified against $X$. 

Our objective is to detect and minimize ungrounded hallucination in $Y_\text{raw}$. Importantly, we do not assume direct access to the generation model and hence do not modify $\mathbfcal{F}(\cdot)$. Instead, we post edit $Y_\text{raw}$ into a refined response $Y_\text{refined}$, such that $Y_\text{refined}$ exhibits reduced hallucination while retaining the essence of $Y_\text{raw}$. 

\section{Methodology}

\begin{figure*}
  \centering
  \includegraphics[width=1\columnwidth]{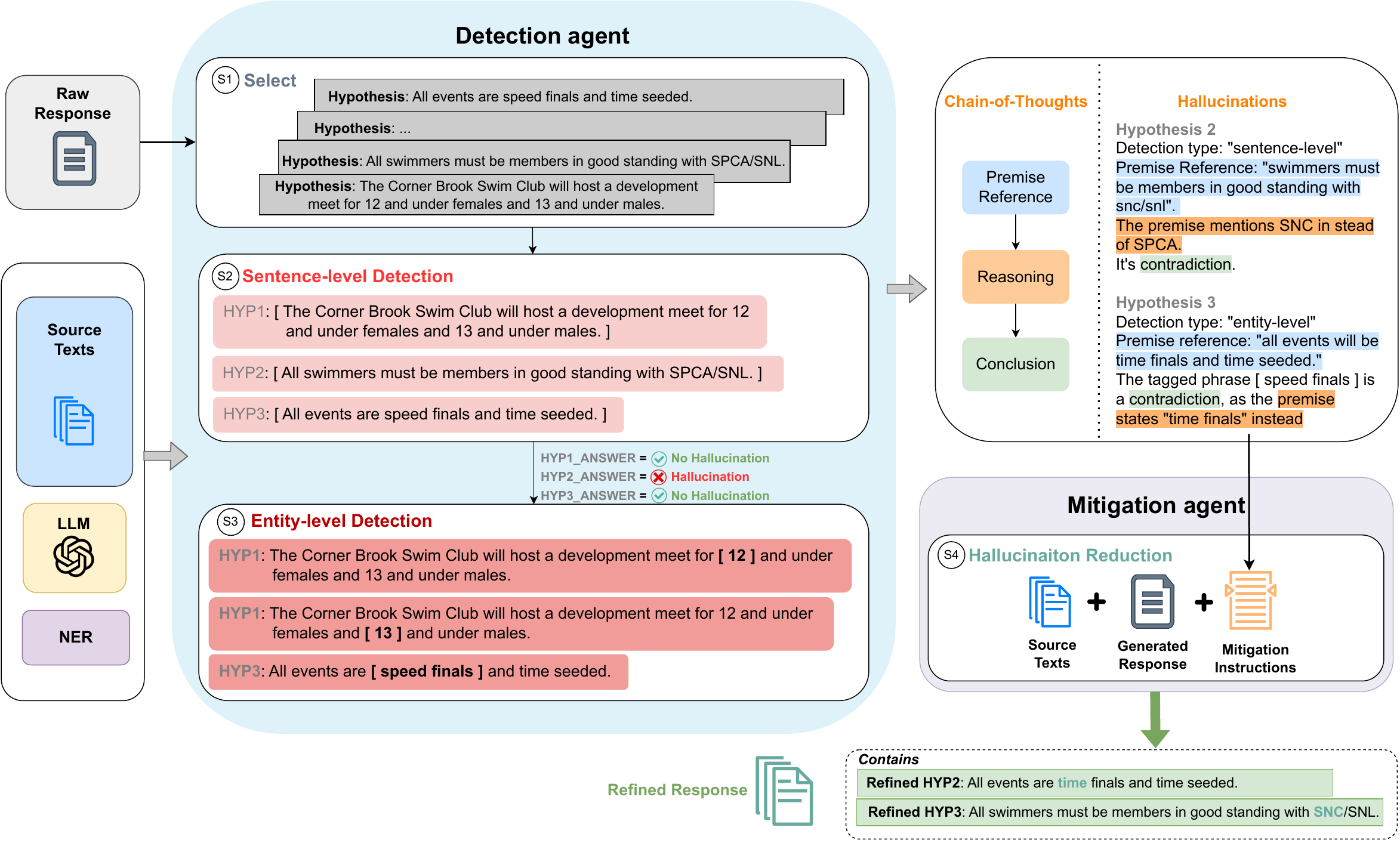}
  \caption{\textbf{Illustration of the proposed framework CoNLI with a real example.} Each hypothesis in the raw response will first go through sentence-level detection. If no hallucination is detected, it will go to detailed entity-level detection. Detection reasonings will be used as mitigation instructions.}~\label{fig:CoNLI}
\end{figure*}

Our solution is a two-stage framework, comprising a \textit{detection agent} and a \textit{mitigation agent} illustrated in Figure \ref{fig:CoNLI} using an example. We provide in-depth discussion of each agent in below sections.

\subsection{Detection agent}
\label{sec:hd}
We formally define $\mathcal{H}_\text{selected} = \{hyp_1, hyp_2,...,hyp_n\}$ as a set of selected hypotheses from $Y_\text{raw}$ for detection; $\mathcal{R} = \{r_1, r_2,...,r_n\}$ as set of reasons against each hypothesis, $\mathcal{J} = \{\text{hallucination, non\_hallucination}\}$ is the final judgement for a hypothesis, further divides into elementary events $\mathcal{J}^+ = \{\text{hallucination}\}$, $\mathcal{J}^- = \{\text{non\_hallucination}\}$. $\mathcal{O}$ is the output of detection agent. Therefore, detection agent can be formulated as:
\begin{equation}
\mathbfcal{D}: (X, Y_\text{raw}) \rightarrow \mathcal{O}\
\end{equation}
\begin{equation}
    \mathcal{O} = \{(hyp_i,r_i, j_i)\}  \subseteq \mathcal{H}_\text{selected} \times \mathcal{R} \times \mathcal{J}
\end{equation}
where we break down $\mathbfcal{D}(\cdot)$ hierarchically into sentence-level detection $\mathbfcal{D}_\text{sent}(\cdot)$ and entity-level detection $\mathbfcal{D}_\text{ent}(\cdot)$ described in below paragraphs. In Addition, given $\mathcal{J}$ is a pair set, this detection phase can be treated as a binary classification. Beyond serving as a precursor to mitigation agent, this module can be independently utilized to evaluate the groundedness of raw response in text-to-text generation applications. Detection agent contains the following steps.



\paragraph{Split and select}
Each raw response $Y_\text{raw}$ is segmented into individual sentences using the NLTK sentence splitter\footnote{\small \url{https://www.nltk.org/api/nltk.tokenize.sent_tokenize.html}}. Sentences that are considered noise or lack factual information for judgement are then purged. For benchmark comparison purposes, we skip this purging process for short-generated responses that can be directly formulated as hypotheses. We leave building advanced hypothesis selector as future work. After this step, we have hypotheses set $\mathcal{H}_\text{selected}$.

\paragraph{Sentence-level detection}
To formulate the NLI problem, we treat the $X$ as the premise for hypotheses $\mathcal{H}$.
The sentence-level detection will sequentially judge each hypothesis independently against the corresponding premise, and categorize them as entailment, contradiction or neutral following \cite{liu2023evaluating}:
\begin{itemize}
    \item \textit{Entailment}: $X \implies hyp_i$
    \item \textit{Contradiction}: $X \implies \neg hyp_i$
    \item \textit{Neutral}: $X \centernot \implies hyp_i$
\end{itemize}

In the ungrounded hallucination scenario, both contradiction and neutral categories in NLI are not aligned with the source, so we treat these two categories as hallucinations. Therefore: 
\begin{equation}
 \mathbfcal{D}_\text{sent}: (X, \mathcal{H}_\text{selected}) \rightarrow \mathcal{O}_\text{sent}
\end{equation}
\begin{equation}
    \mathcal{O}_\text{sent} = \{(hyp_i,r_i^\text{sent}, j_i^\text{sent})\} \subseteq \mathcal{H} \times \mathcal{R}_\text{sent} \times \mathcal{J}
\end{equation}
We divide $\mathcal{O}_\text{sent} = \mathcal{O}_\text{sent}^+ \cup \mathcal{O}_\text{sent}^-$ where hallucination detection output $\mathcal{O}_\text{sent}^+ \subseteq \mathcal{H}_\text{sent}^+ \times \mathcal{R}_\text{sent}^+ \times \mathcal{J}^+$ and non-hallucination detection output $\mathcal{O}_\text{sent}^- \subseteq  \mathcal{H}_\text{sent}^- \times \mathcal{R}_\text{sent}^- \times \mathcal{J}^-$.

We utilize Chain-of-Thought (CoT) prompting \cite{wei2022chain}, guiding the LLM to locate relevant passages in the source text $X$ and allow it to reason and then make a conclusion. To enhance adaptability across domains without intricate prompt engineering, we employ domain-agnostic NLI few-shot examples to orient the LLM towards the essential NLI concepts and the CoT methodology. The specific prompt used in our experiments is detailed in Appendix \ref{sec:DetectionPrompt}. Note that in the few-shot examples, with a given premise, we provide multiple hypotheses and CoT answers in the form of bullet points. This is for batching support so that we may send multiple claims in a single prompt to make our solution more cost-efficient. For benchmarking experiments mentioned in the below sections, we maintain the few-shot examples but disable batching, sending one claim for judgment at a time for apples-to-apples comparison with the other approaches. 

\paragraph{Entity-level detection}
Upon sentence-level evaluation, hypotheses deemed as non-hallucinations undergo subsequent entity-level inspections. This is based on our empirical findings that LLMs, when doing NLI reasonings, may potentially overlook details in the hypothesis and focus more on surface-level semantic features for judgments. If a hypothesis contains abundant factual details or some details require complex reasoning against the source text, sentence-level detection may reach false negative conclusions. Hence, we use entity-level detection to take another look into the non-hallucinated hypothesis $\mathcal{H}_\text{sent}^-$ in $\mathcal{O}_\text{sent}^-$.

Specifically, it will first leverage an entity recognition model (NER) to find entities in the non-hallucinated hypothesis $E = \text{NER}(\mathcal{H}_\text{sent}^-)$. Then it will convert each hypothesis into a sequence of hypothesis where each of them contain a tagged entity to focus on:
\begin{equation}
    \mathbf{f}: \text{hyp}_i \rightarrow \{\text{hyp}_i^e\}, e \in E
\end{equation}

However, unlike $\mathbfcal{D}_\text{sent}$, $\mathbfcal{D}_\text{ent}$ will focus only on the tagged entity without needing to judge other factual information of a hypothesis. This forces the LLM to reason and make judgments against every entities in the non-hallucination hypothesis output by sentence-level detection. If a single $\text{hyp}_i^e \in \text{hyp}_i^E$ is judged as hallucination, we say entity-level judges $\text{hyp}_i$ as hallucination.

\begin{equation}
    \mathbfcal{D}_\text{ent}: (X, \{\text{hyp}_i^e\}) \rightarrow \mathcal{O}_\text{ent}
\end{equation}
\begin{equation}
        \mathcal{O}_\text{ent} =\{(hyp_i,r_i^\text{ent}, j_i^\text{ent})\} \in \mathcal{H}_\text{sent}^- \times \mathcal{R}_\text{ent} \times \mathcal{J}
\end{equation}



\paragraph{Merging}
For each sentence in the generated response, detection agent's final judgment will be $\mathcal{O}=\mathcal{O}_{sent}^+ \cup {O}_{ent}$. For each tuple $\{(hyp_i,r_i, j_i)\}$ in $\mathcal{O}$ where $j_i=\text{hallucination}$, $r_i$ is either a single sentence-level is-hallucination reason or single/multiple entity-level reasons. In other words, a hypothesis will be judged as non-hallucination only if overall sentence judgment and tagged entities judgments all vote for non-hallucination. 


\subsection{Mitigation agent}
Mitigation agent can be formulated $\mathbfcal{M}: (X,Y_\text{raw},\mathcal{O}) \rightarrow Y_\text{refined}$. We consider the hallucination detection result $\mathcal{O}$ as crucial guidance for mitigation agent to reason on how to rewrite these sentences and address issues provided by detection agent. We directly leverage $\mathcal{O}$ as instructions to rewrite $Y_\text{raw}$. Mitigation agent tries to preserve the format of the generated response to the greatest extent possible. It strictly trusts and follows the instructions from detection agent without engaging in additional reasoning on hallucinations. As a result, it could solely focus on how to maintain the fluency and coherency of refined responses by choosing whether to remove or rewrite the hallucination sentences. The prompt used can be found in Appendix \ref{sec:MitigationPrompt}.

\begin{algorithm}[ht]
\small
    \caption{CoNLI hallucination detection and mitigation}\label{alg:example}
	\KwInput{the source text $X$ and the graw response $Y_\text{raw}$ from a text-to-text application}
	\KwOutput{refined response with reduced hallucination $Y_\text{refined}$ }
        \textit{/* Detection agent process*/} \\
        $ \{hyp_1, ...,hyp_n\} =$ HypothesesSelector($Y_\text{raw}$)\;  
	\For{$i=1$ to $n$}
        {
            \If{$hyp_i$ fits the hypothesis selection requirements}
            {
                 ($hyp_i$, $r_i^\text{sent}$, $j_i^\text{sent}$)= $\mathbfcal{D}_\text{sent}$($X$, $\text{hyp}_i$) \;
                 \If{$j_i^\text{sent}$ == non\_hallucinated}
                     {
                          $E = \text{NER}(hyp_i)$ \;
                          \For{$e$ in $E$}
                          {
                                $\mathcal{O}$[$\text{hyp}_i$] += $\mathbfcal{D}_\text{ent}$($X$, $\text{hyp}_i^e$)
                          }
                     }
                 \Else
                    {
                        $\mathcal{O}$[$\text{hyp}_i$] = ($hyp_i$, $r_i^\text{sent}$, $j_i^\text{sent}$)
                    }
            } 
            \Else
            {
                $\mathcal{O}$[$\text{hyp}_i$] = ($hyp_i$, $null$, $non\_hallucination$)
            }
        }
        \textit{/* Mitigation agent process*/} \\
        $Y_\text{refined}$ = Mitigation($X$, $Y_\text{raw}$, $\mathcal{O}$) \\
        \Return  $Y_\text{refined}$
\end{algorithm}

\section{Experiments}
We break down our experiments into two parts. For hallucination detection experiments, we analyze our detection agent's ungrounded hallucination detection performance on various benchmarks and compare it with existing LLM-based and model-based approaches to check our detection quality. For hallucination reduction experiments, we then leverage detection agent's output to do hallucination reduction via mitigation agent on the same benchmarks and do before/after comparisons with text-to-text and hallucination metrics. We try to answer the following two questions:

\smallskip
\noindent
\textit{\textbf{Q1 (Detection)}: How does the performance of our CoNLI detection agent compare to LLM-based and model-based hallucination detection methods?}

\smallskip
\noindent
\textit{\textbf{Q2 (Detection and reduction)}: Does applying CoNLI with hallucination reduction lead to improvements on on NLG and groundedness metrics compared to raw response?
}

\subsection{Hallucination detection experiments}
We conduct experiments on ungrounded hallucination detection with our detection agent.

\subsubsection{Datasets}
We conduct experiments on two different kinds of datasets: (1) datasets with synthetic hallucination generated on ground truth response text. They have larger dataset sizes with defined hallucination categories for easy analysis. (2) datasets with hallucination annotated manually on real state-of-the-art (SOTA) NLG model output response text. They are smaller than the synthetic data, but their hallucinations are closer to hallucinations found in LLM real-world products.

For synthetic datasets, we use a recent LLM hallucination evaluation benchmark HaluEVAL \cite{li2023helma}. We only use summarization and question answering datasets in HaluEval as they contain grounding source texts. We also conducted experiments using annotated datasets traditionally employed for evaluating factual consistency metrics. These datasets include FactCC's summarization test set \cite{kryscinski2020evaluating, cao2020factual}, SummEval \cite{fabbri2021summeval}, QAGS-Xsum \cite{wang2020asking}, QAGS-CNNDM \cite{wang2020asking}. Conventional factual consistency evaluation approaches output consistency scores and use Spearman Correlation coefficients, ROC-AUC \cite{bradley1997use} for evaluation. In our defined groundedness scenario, we consider hallucination as a binary question. Therefore, we use F1 to uniformly evaluate both hallucination evaluation and factual consistency evaluation datasets. We selected a subset of HaluEval benchmark with details mentioned below and factual consistency evaluation datasets we use the same setting following previous works \cite{Zha2023AlignScoreEF, liu2023gpteval}. Dataset statistics can be found in Table~\ref{table:dataset}.


\begin{table}[ht]
\caption{\textbf{Dataset statistics.} We conduct separate experiments on two distinct types of datasets: datasets with synthetic hallucination and substantial dataset size; datasets with hallucination annotated on SOTA NLG model outputs, smaller but closer to application scenarios.}
\label{table:dataset}
\centering
\resizebox{0.99\columnwidth}{!}{
\begin{tabular}{llccc}  
\toprule[1.0pt]
Type & Dataset & Total\# & Hallucination\# & Non\_hallucination\# \\
\midrule[0.3pt]
\multirow{2}{*}{{Synthetic Hallucination}} &
HaluSum2130 \cite{li2023helma} & 2130 & 1065 & 1065 \\
& HaluQA4170 \cite{li2023helma} & 4170 & 2085 & 2085 \\
\midrule[0.3pt]
\multirow{4}{*}{{Annotated Hallucination on SOTA Model Output}} &
FactCC503 \cite{kryscinski2020evaluating} & 503 & 62 & 441 \\
& SummEval \cite{fabbri2021summeval} & 1600 & 294 & 1306 \\
& QAGS-CNNDM \cite{wang2020asking} & 235 & 122 & 113 \\
& QAGS-XSUM \cite{wang2020asking} & 239 & 123 & 116 \\
\bottomrule[1.0pt]
\end{tabular}
}
\end{table}

\paragraph{HaluSum2130} subset of HaluEval \cite{li2023helma} summarization dataset. Each source text contains a pair of hallucination and non-hallucination summaries. For cost concerns of running LLM experiments, we randomly select samples and also filter potentially harmful and sensitive (\ie hate, sexual, violence, self-harm) samples to support the recent trend of building responsible LLM.\footnote{\small \url{https://azure.microsoft.com/en-us/products/ai-services/ai-content-safety}}

\paragraph{HaluQA4170} subset of HaluEval \cite{li2023helma} question answering dataset that each source text also contains a pair of hallucination and non-hallucination answers. Similarly, we do a random sample with content filtering applied. To adapt question answering into our proposed NLI approach, we treat each source text as premise and its associated answer as hypothesis, ignoring the question and answer correctness. That is, an associated answer can still be considered as grounded to the source regardless of the correctness or relevance to the question.

\paragraph{FactCC503} is the FactCC \cite{kryscinski2020evaluating} test set that contains source text and summary sentence pairs. Each summary associated with a source text is generated by SOTA models and then broken down into sentences with poorly generated sentences removed \cite{kryscinski2020evaluating}. Each sentence is annotated as hallucination or non-hallucination.

\paragraph{SummEval and QAGS} SummEval contains 1600 examples built on CNN/Dailymaill \cite{see2017get} with consistency score labeled between 0 to 5. QAGS datasets are built with CNN/Dailymaill \cite{see2017get} (QAGS-CNNDM) and XSUM \cite{narayan2018don} (QAGS-XSUM) respectively with consistency scores between 0 to 1. Unlike past consistency studies, we consider hallucination as a yes or no question for detection and reduction purposes. Therefore, we convert the labels of these datasets into a binary. Only maximum consistency samples are considered as non-hallucination and all the rest are considered as hallucination. All hallucinations are manually annotated on recent SOTA models' outputs.

\subsubsection{Experimental setup}
\paragraph{LLM setup and hyperparameters} We evaluate our framework on OpenAI's \textsc{gpt-3.5-turbo-16k} with max input tokens 16,384 and \textsc{gpt-4-32k} with max input tokens 32,768. We leverage Azure OpenAI ChatGPT API to conduct the experiments.\footnote{\small \url{https://azure.microsoft.com/en-in/products/ai-services/openai-service/}} We set the temperature to 0 to reduce randomness and ensure more deterministic outputs. We set the maximum number of tokens for generation to 4096, top\_p to 0.6, and freq\_penalty and presence\_penalty both to 0.
\label{sec:hd-llm-setup}

\paragraph{Entity detection setup} For the NER in entity-level detection, we leverage Azure Text Analytics (TA) API for entity detection which supports a wide range of entity categories.\footnote{\small \url{https://azure.microsoft.com/en-us/products/ai-services/text-analytics}} Among all the available entity categories, we select the best collection of 9 entities based on the average performance on available validation datasets. Although we observe each experiment dataset has its own best TA categories, to make CoNLI generalizable, we use the same TA categories for all detection and mitigation experiments. See Appendix \ref{sec:TACategory} for more details on the selected TA categories. 



\paragraph{Evaluation metrics} We used F1 since we define our groundedness task as a binary classification. LLM-based hallucination detection approaches usually output binary predictions, while factual consistency evaluation approaches usually output multi-level scores for finer-grained evaluation. Using F1 can unify the measurement for both. We report the macro F1 as well as its breakdowns on hallucination and non-hallucination since the hallucinations can be skewed as per Table~\ref{table:dataset}.

\subsubsection{Results}

\paragraph{Synthetic hallucination dataset results} 
\label{sec:hd-result-synth}
We show the results in Table~\ref{table:HD-result-synthetic}. FactCC and AlignScore are classification models that use alignment output logits as factual consistency scores. We adopt the threshold of 0.5 as the cut-off point for hallucination/non-hallucination predictions, since both are off-the-shelf solutions that aim to be generic with no necessity of downstream fine-tuning. To determine their performance upper-bound, we also investigate their oracle thresholds that best performed on experimented datasets. Notably, the oracle threshold diverges from one dataset to another (see Appendix \ref{sec:threshold}).To establish a unified threshold for generalization, we select the average oracle threshold that yields the highest average F1-macro across all 6 experimented datasets, ensuring a balanced and consistent assessment.

In the case of HaluEval, its provided detection solutions are not task agnostic but designed per their own dataset. Thus we run with its best settings tailored to its own synthetic datasets and skip experiment on annotated hallucination dataset. When running HaluEval, we observed a significant divergence in the behavior of GPT-4 compared to GPT-3.5. GPT-4 exhibited challenges in comprehending the few-shot labels as instructed, resulting in unexpected large performance drops. To mitigate this issue we made an adjustment by appending an additional sentence to the original prompts, which explicitly instructs GPT4 as follows: "for hallucination answer Yes and for non-hallucination answer No". This clarification ensures more accurate performance of HaluEval-GPT4 (*).

We observed that our CoNLI-GPT4 achieves the best F1 on both datasets and averages. It even surpasses AlignScore-Large with upper-bound oracle threshold. Our CoNLI-GPT3.5 achieves the second best averaged F1 and outperforms all listed solutions except those with oracle.

\begin{table}[ht]
\caption{\textbf{Synthetic hallucination dataset results} on F1-macro and breakdown on F1-Hallucination and F1-non\_Hallucination. The last column \textbf{AVG} is the average performance of each metric. Dark green indicates best metric and light green indicates second best on each dataset or average. (*) details addressed in section \ref{sec:hd-result-synth}}
\label{table:HD-result-synthetic}
\centering
\resizebox{0.99\columnwidth}{!}{
\begin{tabular}{lccc|ccc|ccc}  
\toprule[1.0pt]
 & \multicolumn{3}{c}{HaluSum2130} & \multicolumn{3}{c}{HaluQA4170} & \multicolumn{3}{c}{\textbf{AVG}} \\
\midrule[0.3pt]
\textbf{Method} & \textbf{F1} & \textbf{-NonHal} & \textbf{-Hal} & \textbf{F1} & \textbf{-NonHal} & \textbf{-Hal} & \textbf{F1} & \textbf{-NonHal} & \textbf{-Hal} \\
\midrule[0.3pt]
\midrule[0.3pt]
FactCC & 0.421 & 0.224 & 0.618 & 0.485 & 0.412 & 0.558 & 0.453 & 0.318 & 0.588 \\
FactCC (Oracle) & 0.437 & 0.272 & 0.602 & 0.482 & 0.429 & 0.535 & 0.460 & 0.351 & 0.569 \\
AlignScore-L & 0.617 & \cellcolor{lightgreen}0.723 & 0.510 & 0.783 & 0.796 & 0.770 & 0.700 & \cellcolor{lightgreen}0.760 & 0.640 \\
AlignScore-L (Oracle) & \cellcolor{lightgreen}0.669 & 0.665 & \cellcolor{darkgreen}0.672 & 0.750 & 0.743 & 0.756 & 0.710 & 0.704 & 0.714 \\
\midrule[0.3pt]
HaluEval-GPT3.5 & 0.535 & 0.699 & 0.371 & 0.679 & 0.736 & 0.622 & 0.607 & 0.718 & 0.497 \\
HaluEval-GPT4 & 0.415* & 0.683* & 0.147* & 0.796* & 0.827* & 0.764* & 0.606* & 0.755* & 0.456* \\
\midrule[0.3pt]
\textbf{CoNLI-GPT3.5} & 0.633 & 0.641 & 0.624 & \cellcolor{lightgreen}0.848 & \cellcolor{lightgreen} 0.850 & \cellcolor{darkgreen}0.845 & \cellcolor{lightgreen}0.741 & 0.746 & \cellcolor{darkgreen}0.735 \\
\textbf{CoNLI-GPT4} & \cellcolor{darkgreen}0.677 & \cellcolor{darkgreen}0.725 & \cellcolor{lightgreen}0.628 & \cellcolor{darkgreen}0.849 & \cellcolor{darkgreen}0.862 & \cellcolor{lightgreen}0.835 & \cellcolor{darkgreen}0.763 & \cellcolor{darkgreen}0.794 & \cellcolor{lightgreen}0.732 \\
\bottomrule[1.0pt]
\end{tabular}
}
\end{table}

\paragraph{Annotated hallucination dataset results} Shows in Table~\ref{table:HD-result-facteval}. CoNLI-GPT4 achieves the best results on three datasets and averaged, and only underperforms AlignScore-Large averaged with oracle threshold on QAGS-CNNDM. This demonstrates CoNLI, as a generic solution, can achieve high-quality performance in detecting hallucinations in SOTA NLG model outputs. It's also worth mentioning that despite being a much smaller model comparing to GPT-4, AlignScore-Large can also achieve decent performance when an oracle threshold for binary classification is provided. This aligns with its reported high performance on factual consistency evaluation datasets using AUC-ROC and Spearman Correlation coefficients as measurement metrics. Consequently, we think the exploration of finding automatic threshold per task without fine-tuning is an interesting topic for evaluation-score-based approaches. Such study could enhance the applicability of score-based methods to a boarder range of hallucination detection and reduction applications that require a binary answer.

\begin{table}[ht]
\caption{\textbf{Annotated hallucination dataset results} on F1-macro and breakdown on F1-Hallucination and F1-non\_Hallucination. We report their results with classification threshold of 0.5 and of best average across 6 datasets. The last column \textbf{AVG} is the average performance of each metric. }
\label{table:HD-result-facteval}
\centering
\resizebox{0.99\columnwidth}{!}{
\begin{tabular}{l ccc|ccc|ccc|ccc|ccc}  
\toprule[1.0pt]
 & \multicolumn{3}{c}{FactCC503} & \multicolumn{3}{c}{SummEval} & \multicolumn{3}{c}{QAGS-CNNDM}  & \multicolumn{3}{c}{QAGS-XSUM} & \multicolumn{3}{c}{\textbf{AVG}}   \\
\midrule[0.3pt]
\textbf{Method} & \textbf{F1} & \textbf{-NonHal}& \textbf{-Hal} & \textbf{F1} & \textbf{-NonHal}& \textbf{-Hal} & \textbf{F1} & \textbf{-NonHal}& \textbf{-Hal} & \textbf{F1} & \textbf{-NonHal}& \textbf{-Hal} & \textbf{F1} & \textbf{-NonHal}& \textbf{-Hal} \\
\midrule[0.3pt]
\midrule[0.3pt]
FactCC & 0.706 & 0.919 & 0.493 & 0.641 & 0.819 & 0.462 & 0.688 & 0.664 & 0.712 & 0.644 & 0.635 & 0.653 & 0.700 & 0.759 & 0.580 \\
FactCC (Oracle) & 0.710 & 0.923 & 0.496 & 0.651 & 0.833 & 0.468 & 0.689 & 0.678 & 0.700 & 0.649 & 0.653 & 0.644 & 0.674 & 0.772 & 0.577 \\
\midrule[0.3pt]
AlignScore-L & \cellcolor{lightgreen}0.820 & \cellcolor{lightgreen}0.952 & \cellcolor{lightgreen}0.687 & 0.656 & 0.917 & 0.395 & 0.549 & 0.701 & 0.397 & 0.723 & \cellcolor{lightgreen}0.760 & 0.686 & 0.695 & 0.833 & 0.541\\
AlignScore-L (Oracle) & 0.765 & 0.923 & 0.606 & \cellcolor{lightgreen}0.753 & \cellcolor{lightgreen}0.919 & \cellcolor{lightgreen}0.586 & \cellcolor{darkgreen}0.829 & \cellcolor{darkgreen}0.837 & \cellcolor{darkgreen}0.821 & \cellcolor{lightgreen}0.745 & 0.755 & \cellcolor{lightgreen}0.734 & \cellcolor{lightgreen} 0.773 & \cellcolor{lightgreen} 0.859 & \cellcolor{lightgreen} 0.687\\
\midrule[0.3pt]
\textbf{CoNLI-4} & \cellcolor{darkgreen}0.876 & \cellcolor{darkgreen}0.971 & \cellcolor{darkgreen}0.780 & \cellcolor{darkgreen}0.784 & \cellcolor{darkgreen}0.935 & \cellcolor{darkgreen}0.632 &\cellcolor{lightgreen} 0.799 & \cellcolor{lightgreen}0.814 & \cellcolor{lightgreen}0.783 & \cellcolor{darkgreen}0.812 & \cellcolor{darkgreen}0.819 & \cellcolor{darkgreen}0.804 & \cellcolor{darkgreen}0.818 & \cellcolor{darkgreen}0.885 & \cellcolor{darkgreen}0.750 \\
\bottomrule[1.0pt]
\end{tabular}
}  
\end{table}

\paragraph{Ablation study}
We run different variants of CoNLI on the HaluSum2130, HaluQA4170 and FactCC503. Results are presented in Table~\ref{table:HD-ablation}. For entity-detection-only approach, we run entity detection on all hypothesis. For the default hierarchical approach, entity-level detection is only triggered on hypotheses where no hallucination is detected at sentence-level. 

We observe that both sentence-level and entity-level detection results consistently underperform when compared to the combined hierarchical approach. Furthermore, sentence-level results consistently outperform entity-level results, which is logical since entity-level detections within each hypothesis focus solely on tagged entities, whereas sentence-level detection considers the entire hypothesis. Therefore, entity-level detection can be viewed as a valuable augmentation to the sentence-level detector. These findings hold true for both GPT-3.5 and GPT-4 settings.

\begin{table}[ht]
\caption{\textbf{Ablation study for hallucination detection.} We compare CoNLI with sentence-level detection only (sent), entity-level detection only (ent) and hierarchical detection (sent + ent) on GPT3.5 and GPT4.}
\label{table:HD-ablation}
\centering
\resizebox{0.9\columnwidth}{!}{
\begin{tabular}{l ccc|ccc|ccc}  
\toprule[1.0pt]
 & \multicolumn{3}{c}{HaluSum2130} & \multicolumn{3}{c}{HaluQA4170} & \multicolumn{3}{c}{FactCC503} \\
\midrule[0.3pt]
\textbf{Method} & \textbf{F1} & \textbf{-NonHal} & \textbf{-Hal} & \textbf{F1} & \textbf{-NonHal} & \textbf{-Hal} & \textbf{F1} & \textbf{-NonHal} & \textbf{-Hal}  \\
\midrule[0.3pt]
\midrule[0.3pt]
CoNLI-3.5 (sent)  & 0.628 & \textbf{0.737} & 0.519 & 0.809 & 0.824 & 0.793 & 0.668 & 0.931 & 0.404 \\
CoNLI-3.5 (ent) & 0.647 & 0.692 & 0.601 & 0.783 & 0.820 & 0.745 & 0.652 & 0.909 & 0.394 \\
\cellcolor{na}CoNLI-3.5 (sent+ent) & \textbf{0.664} & 0.695 & \textbf{0.632}  &  \textbf{0.840} & \textbf{0.845} & \textbf{0.834} & \textbf{0.694} & \textbf{0.933} & \textbf{0.455} \\
\midrule[0.3pt]
CoNLI-4 (sent) & 0.666 & \textbf{0.753} & 0.578 
 & 0.832 & 0.850 & 0.813 & 0.858 & 0.968 & 0.748 \\
CoNLI-4 (ent) & 0.667 & 0.738 & 0.595 & 0.771 & 0.817 & 0.724  & 0.834 & 0.964 & 0.704\\
\cellcolor{na}CoNLI-4 (sent+ent) &  \textbf{0.677} & 0.725 & \textbf{0.628} & \textbf{0.844} & \textbf{0.859} & 
\textbf{0.829} & \textbf{0.876} & \textbf{0.971} & \textbf{0.780} \\
\bottomrule[1.0pt]
\end{tabular}
}
\end{table}

\subsection{Hallucination reduction experiments}
In this section, we conduct experiments on evaluating our CoNLI performance end-to-end with detection agent and mitigation agent combined. We used the same LLM setup and hyperparameters as detection expeirment mentioned in section \ref{sec:hd-llm-setup}. 

\subsubsection{Experimental setup}
\paragraph{Datasets} As a subsequent experiment in the context of hallucination detection detection, we continue to use HaluSum2130, HaluQA4170 synthetic datasets to experiments at larger scale. Additionally, we incorporate the human-annotated FactCC503 dataset, which encompasses hallucinations from a diverse set of 10 SOTA NLG models,  making it the most comprehensive among the annotated hallucination datasets mentioned.

For HaluSum2130 and HaluQA4170, we use the non-hallucination summary as the ground truth for non-hallucination summaries. In the case of the FactCC503, we aggregate sentence-level summarization data into comprehensive summary. Subsequently, we apply our detection agent judgment on a per sentence basis to refine the complete summary and compare to the ground truth summary. 


\paragraph{Evaluation metrics} We evaluate text response quality in conventional NLG metrics Rouge1, Rouge2, RougeL, Bleu-4, BertScore \cite{zhang2019bertscore} and hallucination evaluation metrics FactCC \cite{kryscinski2020evaluating} and AlignScore-Large \cite{Zha2023AlignScoreEF}. Furthermore, We use our proposed CoNLI-GPT4 for hallucination evaluation, leveraging its demonstrated high quality in the preceding hallucination detection experiments. For each dataset, the CoNLI-GPT4 score demonstrates the percentage of refined responses containing zero ungrounded hallucination by its detection.


\subsubsection{Results}
We show the hallucination reduction results with before and after CoNLI applied in Table~\ref{table:HM-result}. For synthetic datasets, HaluSum2130 and HaluQA4170, all metrics improved with CoNLI refined response. Responses in question answering datasets are shorter compared to those in summarization datasets. As a result, minor refinements have a more pronounced impact on the evaluation metrics.

In the annotated dataset, FactCC503, we observed a distinct pattern. Given that the raw responses are selected from state-of-the-art NLG models trained to optimize NLG metrics, especially Rouge scores, we noticed a slight decline in Rouge scores after the refinement process. However, it's important to note that this decline in Rouge scores does not necessarily indicate a drop in response quality, because we also observed improvements in BertScore and Bleu score. As Rouge score is more recall focused (\ie amount of n-grams in reference appears in generated response) and Bleu score is more precision focused (\ie amount of n-grams in generated response appears in reference), Bleu score improvement means irrelevant tokens in responses are reduced, indicating a reduction in hallucinatory content. This hypothesis aligned with the consistent improvement on hallucination evaluation metrics, FactCC, AlignScore-Large and CoNLI-GPT4. Therefore, our CoNLI refinement process maintains response quality while effectively reducing hallucinations in the outputs of SOTA NLG models.

\begin{table}[ht]
\caption{\textbf{Hallucination reduction result.} We compare CoNLI refined response with raw generated response on various NLG and hallucination metrics.}
\label{table:HM-result}
\centering
\resizebox{0.99\columnwidth}{!}{
\begin{tabular}{ll cccc c | ccc}
\toprule[1.0pt]
Dataset & Target & Rouge1 & Rouge2 & RougeL & Bleu-4 & BertScore & FactCC & AlignScore-L & CoNLI-4 \\
\midrule[0.3pt]
\multirow{2}{*}{{HaluSum2130}}
 & Raw Response & 38.45 & 14.06 & 34.41 & 8.52 & 88.21 & 18.02 & 57.54  & 49.48 \\
 & Refined Response & \textbf{39.62} & \textbf{15.22} & \textbf{35.54} & \textbf{9.69} & \textbf{88.46} & \textbf{20.66} & \textbf{76.07} & \textbf{66.01} \\
\midrule[0.3pt]
\multirow{2}{*}{{HaluQA4170}}
 & Raw Response & 9.27 & 3.13 & 9.02 & 1.17 & 82.25 & 36.37 & 28.36  & 24.12 \\
 & Refined Response & \textbf{25.48} & \textbf{14.54} & \textbf{25.38} & \textbf{3.98} & \textbf{84.61} & \textbf{40.48} & \textbf{76.21} & \textbf{80.19} \\
\midrule[0.3pt]
\multirow{2}{*}{{FactCC503}}
 & Raw Response & \textbf{31.71} & \textbf{12.02} & \textbf{28.84} & 5.61 & 85.20 & 84.49 & 81.95 & 88.27 \\
 & Refined Response & 31.27 & 11.94 & 28.36 & \textbf{6.12} & \textbf{85.58} & \textbf{87.81} & \textbf{90.83} & \textbf{96.22} \\
\bottomrule[1.0pt]
\end{tabular}
}
\end{table}


\section{Related work}
Hallucination is a well-known issue for text-to-text models \cite{maynez2020faithfulness} including LLM \cite{zhang2023language,mckenna2023sources} and it is a critical problem to apply LLM to real-world applications responsibly. Various recent surveys offers comprehensive examination about this topic \cite{zhang2023siren,rawte2023survey,ji2023survey}.

\paragraph{Hallucinations detection}
Many recent studies focus on evaluating factual consistency, similar scenario as hallucination detection, except they provide consistency score to measure the alignment against grounding source instead of binary prediction of is content hallucination or not. FactCC \cite{kryscinski2020evaluating} leverages foundation language models with generated weakly-supervised training data to train a classification model; Zhou et al. propose token-level hallucination detection and leverage more fine grained
losses to improve quality \cite{zhou2021detecting}; AlignScore \cite{Zha2023AlignScoreEF} develop a unified training framework of the alignment function by integrating a large diversity of data sources. In LLM based approaches, SelfCheckGPT \cite{manakul2023selfcheckgpt} leverages self-consistency of LLM to detect hallucination in runetime by generatimg multiple samples; G-Eval leverages GPT to provide NLG metrics that include factual consistency evaluation \cite{liu2023gpteval}. HaluEval \cite{li2023helma} provides LLM hallucination benchmark on multiple domains supporting grounded and ungrounded hallucination detection. It also proposes an LLM solution leveraging GPT with CoT.

\paragraph{Hallucinations reduction}
In addition to hallucination detection, there is also a growing body of research dedicated to reducing the occurrence of hallucinations in the generated text. ChatProtect detects and mitigates self-conflicting hallucinations in LLM-generated text \cite{mundler2023self}. CoVe \cite{dhuliawala2023chain} reduces hallucination through a sequence of fact verification questions. Moreover, hallucination can be reduced when the LLM that generates response is fully accessible for runtime mitigation  \cite{chuang2023dola, press2022measuring, manakul2023selfcheckgpt, du2023improving} or with the help of external knowledge \cite{gaoetal2023rarr}.

\section{Conclusion}
In this work, we explore how to leverage LLM to efficiently detect and reduce ungrounded hallucinations in a plug-and-play manner. We conduct extensive experiments on a range of text-to-text datasets, addressing both hallucination detection and reduction. We propose a simple yet effective LLM-based framework that formulates hallucination detection into a chain of NLI tasks. It incorporates both sentence-level and entity-level judgements with demonstrated effectiveness. Importantly, its interpretable output can also be leveraged for hallucination reduction. Overall, Our framework's generalizability allows seamless deployment without adjustments and has demonstrated remarkable detection quality and reduced hallucination while preserving text quality.

\section*{Acknowledgement}
We would like to thank all Microsoft Responsible AI team members working on hallucination detection and mitigation efforts. \textbf{Alex Gorevski} for various engineering support; \textbf{Kaushik Chakrabati} for Microsoft internal dataset construction; \textbf{Aaron Aspinwall} for Microsoft internal synthetic dataset construction and for providing valuable review and feedback on the paper; \textbf{Karim Zakaria}, \textbf{Hossam Emam}, \textbf{Wentao Hu} and \textbf{Hongliang Kong} for their contribution to engineering and infrasturcture; \textbf{Aya Shakerm}, \textbf{Yousra Hesham} for their work on science foundations. \textbf{Dan Iter} for providing hallucination mitigation baseline.

\bibliography{main}
\bibliographystyle{unsrt}

\clearpage

\appendix
\section{Hallucinaton Category}
Our work categorizes hallucination into the following categories and subcategories:
\begin{itemize}
    \item Context-free hallucination
    \item Ungrounded hallucination
    \item Self-conflicting hallucination
\end{itemize}
Among all categories, we picked ungrounded hallucination as the focus of our research. We will demonstrate examples for each category and subcategory. 

\begin{figure*}[ht]
  \centering
  \includegraphics[width=0.8\columnwidth]{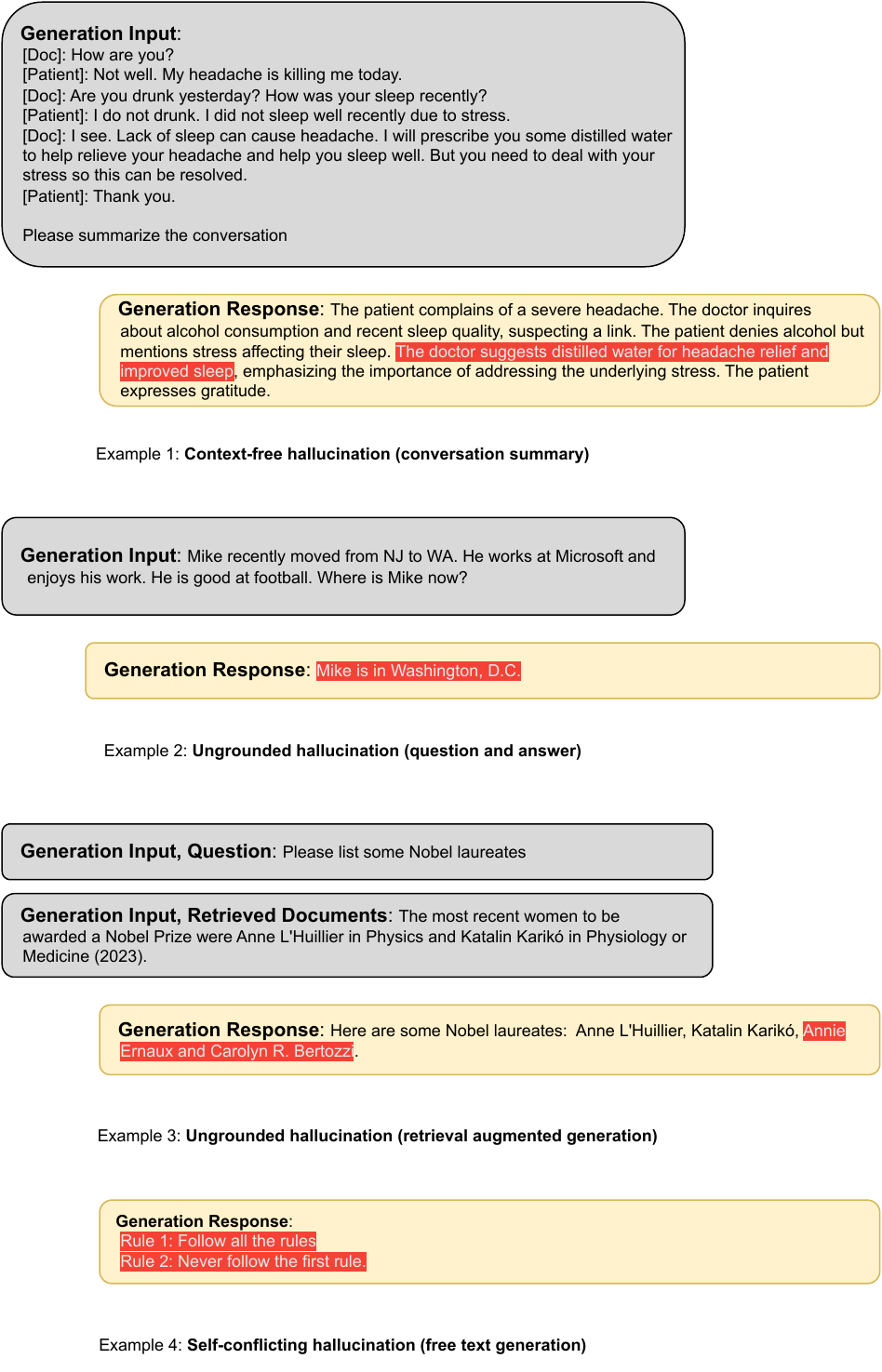}
  \caption{\textbf{Hallucination examples}}~\label{fig:example}
\end{figure*}

Figure \ref{fig:example} shows multiple examples of hallucination:

Example 1 is a context-free hallucination in the conversation summary scenario. Even \textit{"The doctor suggests distilled water for headache relief and improved sleep"} in the summary can be related to \textit{"I will prescribe you some distilled water to help relieve your headache and help sleep well"} in generation input, it contradicts with commonsense and should, therefore, be considered as a context-free hallucination.

Example 2 presents another example with an ungrounded hallucination in a question answer scenario. \textit{"Washington, D.C"} in the generated response contradicts with \textit{"WA"} in the generation input as "WA" should reference to "the Washington state".

Example 3 illustrates another ungrounded hallucination in retrieval augmented generation scenario. There is no source in the generation input to support \textit{"Annie Ernaux and Carolyn R. Bertozzi."} in the generated response, even though it matches commonsense. 

Example 4 illustrates a self-conflicting hallucination in a free text generation scenario. In the given example, the first rule contradicts the second rule.

\section{Entity category definition}
There are a total of 37 different entities leveraging TA among which we picked 9 of them:\footnote{\small \url{https://learn.microsoft.com/en-us/azure/ai-services/language-service/named-entity-recognition/concepts/named-entity-categories?tabs=ga-api}}
\begin{itemize}
\item\textsc{Person}: Names of people.
\item\textsc{PersonType}: Job types or roles held by a person
\item\textsc{Location}: Natural and human-made landmarks, structures, geographical features, and geopolitical entities.
\item\textsc{Event}: Historical, social, and naturally occurring events.
\item\textsc{Skill}: A capability, skill, or expertise.
\item\textsc{DateTime-DateRange}: Date ranges.
\item\textsc{DateTime-Duration}: Durations.
\item\textsc{Quantity-Number}: Numbers.
\item\textsc{Quantity-Currency}: Currencies
\end{itemize}
\label{sec:TACategory}

\section{FactCC and AlignScore threshold on datasets}
\label{sec:threshold}
In our experiment, we noted that the optimal thresholds for FactCC and AlignScore-Large vary considerably across different datasets. This variability poses a challenge in selecting a uniform threshold for all available datasets. Consequently, we decided to report the threshold that produced the highest average F1-macro score across all 6 datasets. For further specifics, refer to Table~\ref{table:thresholds-on-datasets}.

\begin{table}[ht!]
\caption{FactCC and AlignScore optimal threshold on each dataset and the threshold that yields the best average across all available datasets}
\label{table:thresholds-on-datasets}
\centering
\resizebox{0.99\columnwidth}{!}{
\begin{tabular}{lc|c|c|c|c|c|c}  
\toprule[1.0pt]
 & HaluSum2130 & HaluQA4170 & FactCC503 & SummEval & QAGS-CNNDM  & QAGS-XSUM & AVG \\
\midrule[0.3pt]
FactCC & 0.02 & 0.94 & 0.90 & 0.14 & 0.38 & 0.24 & \textbf{0.14} \\
AlignScore-L & 0.66 & 0.38 & 0.14 & 0.80 & 0.80 & 0.94  & \textbf{0.74}  \\
\bottomrule[1.0pt]
\end{tabular}
}
\end{table}

\section{Detection agent prompt}
\label{sec:DetectionPrompt}
The detection prompt can be divided into sections of system information, first few-shot example, second few-shot example, and raw response.
\subsection{System instruction}

 \textit{You are a helpful assistant. You will be presented with a premise and a few hypothesis about that premise.} \\
 \\
 \textit{A hypothesis is usually in forms of a sentence.} \\
 \\
 \textit{A premise is usually a long source document or transcript.} \\
 \\
 \textit{You need to decide whether the hypothesis is entailed by the premise by choosing one of the following:}
 \begin{enumerate}
\item \textit{Entailment: The hypothesis follows logically from the information contained in the premise. Mark \textsc{[C]}.}
\item \textit{Contradiction: The hypothesis is logically false from the information contained in the premise. Mark \textsc{[I]}.}
\item \textit{Neutral: It is not possible to determine whether the hypothesis is true or false without further information. Mark \textsc{[I]}.}
\end{enumerate}
\textit{Read the passage of information thoroughly and select the correct answer either \textsc{[C]} or \textsc{[I]}. Read the premise thoroughly to ensure you know what the premise entails.} \\ 
\textit{For each judgement, think step by step with following guidelines:}
\begin{enumerate}
    \item \textit{Repeat hypothesis you are judging.}
    \item \textit{Find the part of the premise that is related to the hypothesis. If we can not find any, it is not factually correct and thus should be marked as \textsc{[I]}.}
    \item \textit{If we found related part in the premise but it is factually not aligned with the hypothesis, we also mark \textsc{[I]}. If it is factually aligned, we mark it \textsc{[C]}.}
\end{enumerate}
\textit{Try your best to give the right answer.} \\
\textit{Rules:}
\begin{itemize}[label=$\star$]
    \item  \textit{You may assume that today is March 24th, 2023. Use this date when analyzing dates and time spans.}
    \item \textit{Please ignore the age when judging entailment.  If the age is incorrect, and everything else is correct, it is still a factually correct hypothesis that should be marked \textsc{[C]}.}
    \item \textit{If the hypothesis only has less than 3 words with no context, mark \textsc{[C]} if you can find those words in the premise without investigating the context.}
    \item \textit{Please only do reasoning and judge the factual correctness of the hypothesis only. Ignore syntax related issues.}
    \item \textit{Make sure your logic and reasoning are rigorous, intelligent, and defensible.}
    \item \textit{Make conservative judgement. Only mark \textsc{[I]} when you are very confident.}
    \item \textit{Write down the reasoning process first, and then make final conclusion. Mark each hypothesis either \textsc{[C]} or \textsc{[I]}.}
    \item \textit{Write the answers with the original hypothesis ordered format.}
\end{itemize}

\subsection{First few shot example}
\subsubsection{Example user} 
\textit{Let's try it.  I'll give you a premise and a few hypothesis.}\\
\textit{Premise:}\\
\begin{itemize}[label={}]
    \item \textit{The Academy Awards, also known as the Oscars are awards for artistic and technical merit for the film industry. They are presented annually by the Academy of Motion Picture Arts and Sciences, in recognition of excellence in cinematic achievements as assessed by the Academy's voting membership. The Academy Awards are regarded by many as the most prestigious, significant awards in the entertainment industry in the United States and worldwide.} \\
\end{itemize}

\textit{Hypothesis:} \\
\begin{enumerate}[label=(\arabic*).]
    \setcounter{enumi}{-1}
    \item \textit{Oscar is presented every other two years.} \\
    \item \textit{Oscar is very important awards in the entertainment industry in the United States. And it's also significant worldwide.} \\
    \item \textit{Will Smith won the 2022 Oscar.}
\end{enumerate}

\subsubsection{Example assistant}

\textit{Answer:}
\begin{enumerate}[label=(\arabic*).]
    \setcounter{enumi}{-1}
    \item \textit{Oscar is presented every other two years. <reason> premise reference: "They are presented annually by the Academy of Motion Picture Arts and Sciences", thus it is not presented every other two year. It's contradiction. </reason> \textsc{[I]}} \\
    \item \textit{Oscar is very important awards in the entertainment industry in the United States. And it's also significant worldwide. <reason> premise reference: "The Academy Awards are regarded by many as the most prestigious, significant awards in the entertainment industry in the United States and worldwide." It's entailment." </reason> \textsc{[C]}} \\
    \item \textit{Will Smith won the 2022 Oscar.<reason> premise reference: None. The transcript didn't mention information related to Will Smith winning Oscar. It's neutral. </reason> \textsc{[I]}}
\end{enumerate}

\subsection{Second few shot example}
\subsubsection{Example user}
\textit{Let's try it again.  I'll give you a new premise unrelated to previous examples and a few hypothesis. This time there are some tagged [ words ] in the hypothesis. Make judgement and reasoning focusing on the tagged [ words ].} \\
        
\textit{Premise:} \\
\begin{itemize}[label={}]
    \item \textit{In Quebec, an allophone is a resident, usually an immigrant, whose mother tongue or home language is neither French nor English.} \\
\end{itemize}

\textit{Hypothesis:} \\

\begin{enumerate}[label=(\arabic*).]
    \setcounter{enumi}{-1}
    \item \textit{In Quebec, an allophone is a resident, usually an [ immigrant ], whose mother tongue or home language is not French.} \\
    \item \textit{In Quebec, an allophone is a resident whose home language is not [ French ].} \\
    \item \textit{In Quebec, an [ foreigner ] is a resident whose home language is not French.}
\end{enumerate}

\subsubsection{Example assistant} 

\textit{Answer:}
\begin{enumerate}[label=(\arabic*).]
    \setcounter{enumi}{-1}
    \item \textit{In Quebec, an allophone is a resident, usually an [ immigrant ], whose mother tongue or home language is not French. <reason> premise reference: "In Quebec, an allophone is a resident, usually an immigrant." Thus the tagged word [ immigrant ] is an entailment. </reason> \textsc{[C]}} \\
    \item \textit{In Quebec, an allophone is a resident whose home language is not [ French ] <reason> premise reference: "an allophone is a resident, usually an immigrant, whose mother tongue or home language is neither French nor English." French is a subset of "French nor English". The tagged word [ French ] is an entailment. </reason> \textsc{[C]}} \\
    \item \textit{In Quebec, an [ foreigner ] is a resident whose home language is not French. <reason> premise refernece: "an allophone is a resident, usually an immigrant, whose mother tongue or home language is neither French nor English." The premise talks about allophone not foreigner. The tagged word [ allophone ] is an contradiction. </reason> \textsc{[I]}}
\end{enumerate}

\subsection{Current request}
\textit{Now let's try one more time.} \\
\textit{I'll give you a new and unique premise and the previous examples do not apply. I'll also give you a few new hypothesis about the premise.}
\textit{Use all of the instructions given above follow the exact format as above examples to judge each hypothesis. Whether it's contradiction, entailment or neutral, and mark them as either \textsc{[C]} or \textsc{[I]}}

\textit{Premise:} 
\begin{itemize}[label={}]
    \item \textit{\{\{Source Text\}\}}
\end{itemize}

\textit{Hypothesis:}
\begin{itemize}[label={}]
    \item \textit{\{\{Hypothesis\}\}}
\end{itemize}

\textit{Begin your answer with "Answer:\textbackslash
n"}

\section{Mitigation agent prompt}
\label{sec:MitigationPrompt}
\subsection{System instruction}
\textit{You are a proof-reading assistant for a documentation scribe.}  \\

\textit{Given the source DOCUMENT information, the scribe is expected to write factually correct CLAIM for the source using a specified format.} \\

\textit{Read the following DOCUMENT along with the resulting CLAIM and rewrite the CLAIM to correct any discrepancies between the DOCUMENT and CLAIM based on provided instructions.}

\textit{The CLAIM occasionally has errors. Below we provide a list of sentences from the CLAIM that need to be rewritten and why they have issues. All sentences in the CLAIM must be supported by evidence in the DOCUMENT.} \\

\subsection{Current request}

\textit{DOCUMENT:}
\textit{Hypothesis:}
\begin{itemize}[label={}]
    \item \{\{Source Text\}\}
\end{itemize}

\textit{End DOCUMENT.} \\

\textit{CLAIM:}
\begin{itemize}[label={}]
    \item \{\{Raw Response\}\}
\end{itemize}
\textit{End CLAIM.} \\

\textit{Rerwrite these sentences with instructions to the CLAIM:}
\begin{itemize}[label={}]
    \item \{\{Rewrite Instructions\}\} \\
\end{itemize}
\textit{Directly rewrite the CLAIM exactly as it is written above but rewrite the above sentences in the instructions base on the reasons why they are incorrect. Keep the rest sentences unchanged.} 

\textit{For the sentences in above instructions are hard to be rewritten due to no enough information provided in source document, remove theose sentences in the corrected CLAIM.} \\

\textit{Corrected WHOLE CLAIM:}

\textit{Begin your answer with "Answer:\textbackslash n"}


\end{document}